\def\year{2020}\relax
\documentclass[letterpaper]{article} 
\usepackage{aaai20}  
\usepackage{times}  
\usepackage{helvet} 
\usepackage{courier}  
\usepackage[hyphens]{url}  
\usepackage{graphicx} 
\usepackage{graphicx}  
\usepackage{amsmath,amssymb}

\title{Channel Interaction Networks for Fine-Grained Image Categorization}
\author{\Large \textbf{Yu Gao, Xintong Han, Xun Wang, Weilin Huang\thanks{Weilin Huang is the corresponding author.}, Matthew R. Scott
}\\ 
Malong Technologies, Shenzhen, China\\
Shenzhen Malong Artificial Intelligence Research Center, Shenzhen, China\\ 
\{chrgao,xinhan,xunwang,whuang,mscott\}@malong.com 
}

\begin{document}

\maketitle

\begin{abstract}
Fine-grained image categorization is challenging due to the subtle inter-class differences.
We posit that exploiting the rich relationships between channels can help capture such differences since different channels correspond to different semantics.
In this paper, we propose a channel interaction network (CIN), which models the channel-wise interplay both within an image and across images.
For a single image, a self-channel interaction (SCI) module is proposed to explore channel-wise correlation within the image.
This allows the model to learn the complementary features from the correlated channels, yielding stronger fine-grained features.
Furthermore, given an image pair, we introduce a contrastive channel interaction (CCI) module to model the cross-sample channel interaction with a metric learning framework, allowing the CIN to distinguish the subtle visual differences between images. Our model can be trained efficiently in an end-to-end fashion without the need of multi-stage training and testing.
Finally, comprehensive experiments are conducted on three publicly available benchmarks, where the proposed method consistently outperforms the state-of-the-art approaches, such as DFL-CNN\cite{wang2018learning} and NTS\cite{yang2018learning}.
\end{abstract}

\section{Introduction}
\label{sec:intro}
\begin{figure}[t]
\begin{center}
\includegraphics[width=0.48\textwidth]{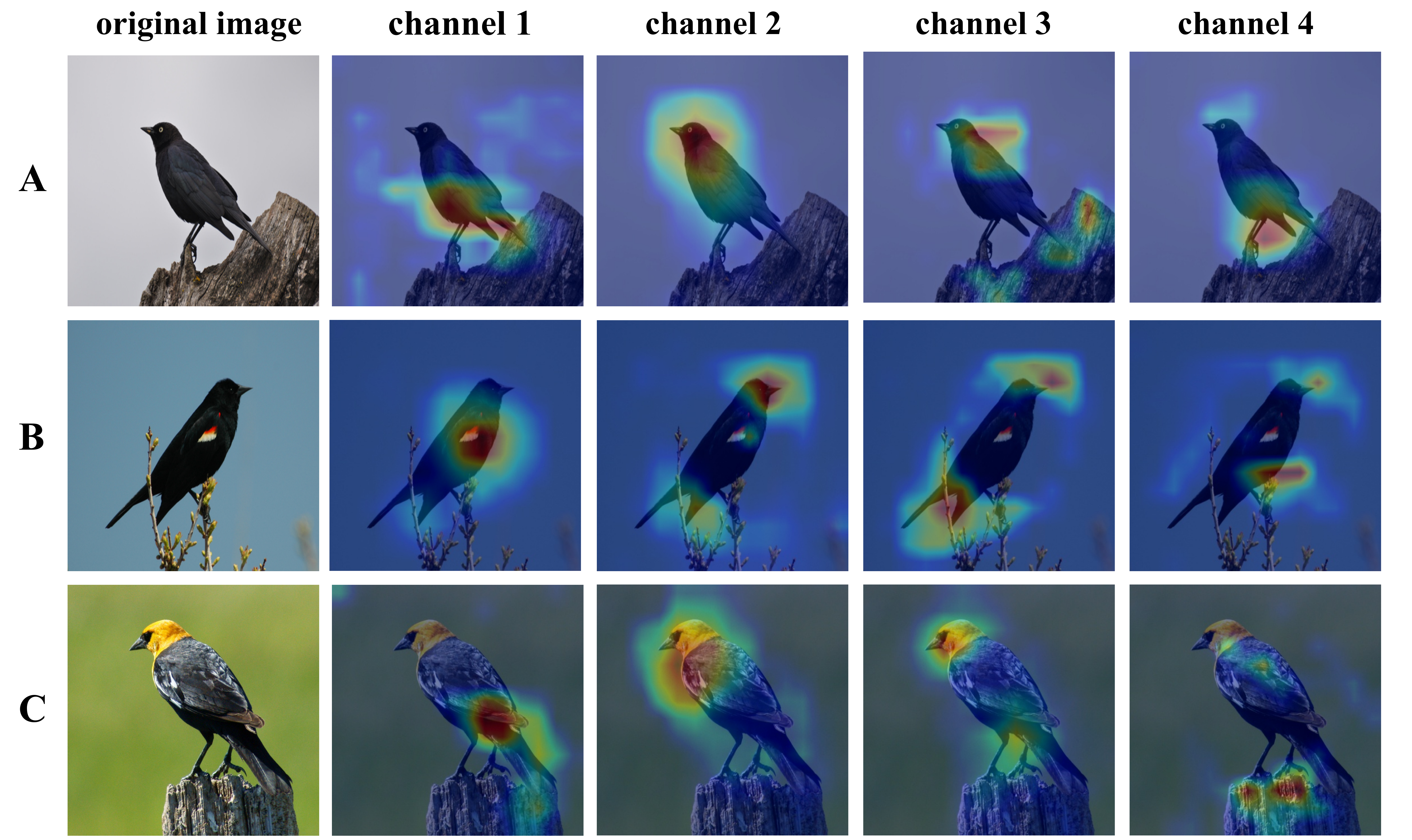}
\end{center}
\caption{Channel activations computed by our method (from the \texttt{conv5\_3} layer of ResNet50 trained on CUB-200-2011 dataset).}
\label{fig:introduction}
\end{figure}

Fine-grained image categorization has become an important topic in computer vision community with broad application prospects such as new retail~\cite{karlinsky2017fine}, automatic driving~\cite{DBLP:conf/cvpr/SochorHH16}, etc. Going beyond classical image classification that recognizes basic-level categories, fine-grained categories are much more challenging to be identified due to the subtle inter-class differences, many of which can only be effectively distinguished by concentrating on discriminative local parts. For instance, to distinguish three bird species in Figure \ref{fig:introduction}, a neural network usually focuses on their wings and heads.

Previous work tends to learn discriminative features by locating distinct parts~\cite{DBLP:conf/nips/JaderbergSZK15,fu2017look,yang2018learning} or modeling higher order information~\cite{lin2015bilinear,DBLP:conf/cvpr/GaoBZD16,DBLP:conf/cvpr/KongF17,yu2018hierarchical}, which have been proven to be effective for fine-grained image classification.
In this paper, we rethink the way of learning discriminative features with convolutional networks, and propose a new channel interaction network (CIN).

First, different channels often correspond to different visual patterns~\cite{yosinski2015understanding}. As shown in Figure \ref{fig:introduction}, most of the channels are semantically complementary to each other. Motivated by this observation, we aim to discover the complementary channel information for each individual channel, and then aggregate the complementary channels with the original ones. Such complementary information can cooperatively contribute to the referred channel, making the channel more discriminative. Consequently, we propose a self-channel interaction (SCI) network that explicitly models the relationships between various channels to discover such channel-wise complementary clues.
Existing methods usually apply the channel/part interplay for direct classification~\cite{lin2015bilinear,yu2018hierarchical} or attempt to mine the closely related cues~\cite{DBLP:journals/corr/abs-1711-07971,yue2018compact}, where the channel-wise complementary clues are not fully explored.

Second, people tend to distinguish two images by focusing on their specific distinctions. For instance, when we compare two images, A and B, in Figure \ref{fig:introduction}, it is easy to identify the difference of wings between two images: the bird in A has black wings while there is  a red dot on the wings of B.
%
However, when images of A and C are compared, more attention should be paid to the regions of heads, i.e., enhancing the importance of channel 2 for image A and image C. To this end, we propose a novel contrastive channel interaction (CCI) mechanism between samples, with the goal of capturing the subtle differences between images. Metric learning is incorporated with our framework to model the cross-sample channel interactions, which is neglected by most of the existing methods~\cite{wang2018learning,yang2018learning}.

\begin{figure*}
\begin{center}
\includegraphics[width=0.95\textwidth]{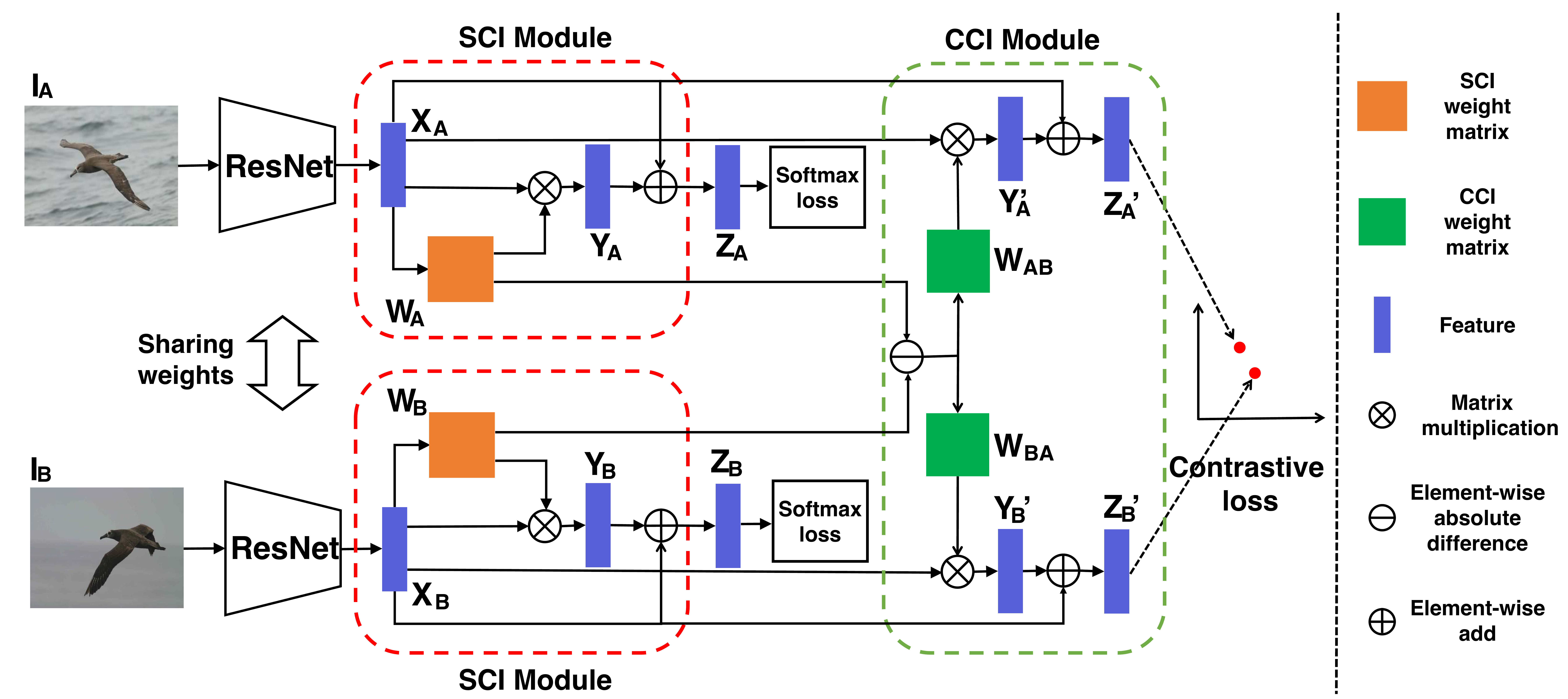}
\end{center}
\caption{
Overview of our network architecture. $W_{AB} = |W_A - \eta W_B|, W_{BA} = |W_B - \gamma W_A|$, $\eta$ and $\gamma$ are learned by an fc layer controlling the encoded information computed from the contrastive image for highlighting differences. CCI module will be removed and softmax loss will be replaced by a softmax layer during inference.}
\label{approach:whole}
\end{figure*}

Finally, we jointly optimize the SCI module and the CCI module, as shown in Figure \ref{approach:whole}. The network can be trained end-to-end in one stage, and thus is more lightweight than the two-stage methods like HS-Net~\cite{lam2017fine}, DFL-CNN~\cite{wang2018learning}, NTS~\cite{yang2018learning}, etc..
Our major contributions are summarized as:

1) We propose a self-channel interaction (SCI) module able to model the interplay between different channels within an image. This enables it to capture the channel-wise complementary  information for each channel, which enhances the discriminative features learned by each channel. This results in a lightweight model that can be trained more effectively in one stage. The new model is flexible, and can be seamlessly integrated into existing networks to boost the performance.

2) We propose a novel contrastive channel interaction (CCI) module to learn channel-wise relationships between images. CCI is able to dynamically identify the distinct regions from two compared images, allowing the model to focus on such distinctive regions for better categorization.

3) Finally, we evaluate our approach on three publicly available datasets: CUB-200-2011 \cite{wah2011caltech}, Stanford Cars \cite{krause20133d} and FGVC Aircraft \cite{DBLP:journals/corr/MajiRKBV13}, where our method achieves better performance over current state-of-the-art.

\section{Related Work}

\textbf{Fine-grained feature representations.} Building powerful feature representations has been broadly studied for fine-grained image categorization.
Unlike using the first-order features directly for classification ~\cite{zhang2014part,DBLP:journals/pr/WeiXWS18},  \cite{lin2015bilinear} employ bilinear pooling with two independent CNNs, which take pairwise feature interactions into consideration by computing the second-order information, leading  to performance improvements.
To reduce the computation complexity,  ~\cite{DBLP:conf/cvpr/GaoBZD16,DBLP:conf/cvpr/KongF17} tend to use less feature dimensions,
while \cite{DBLP:conf/cvpr/CuiZWLLB17} attempt to model higher-order information to improve the accuracy. ~\cite{DBLP:conf/cvpr/WangLZ17,DBLP:conf/cvpr/LiXWG18} apply a matrix power normalization for computing bilinear features.
~\cite{yu2018hierarchical} further explore cross-layer bilinear pooling to compute multi-layer knowledge.
Unlike these methods using second or higher order information for direct classification, we compute second-order statistics between different channels, which are used jointly with the original features to capture the channel-wise complementary information, resulting in stronger deep representations.

\textbf{Visual attention.}
Visual attention, which has been introduced in various  computer vision applications, can be employed to capture the subtle inter-class differences in fine-grained image categorization.
For example, hard attention based methods, such as~\cite{DBLP:conf/nips/JaderbergSZK15,fu2017look,li2017dynamic,yang2018learning}, usually detect local regions and then crop them out from the original image.
But the main limitation is that each cropped region requires an extra feedforward operation.
Instead, soft attention methods~\cite{zheng2017learning,DBLP:conf/eccv/SunYZD18} can be regarded as imposing a soft mask on the feature maps, by only using a single feedforward stage. Self-attention was proposed and applied in machine translate in~\cite{vaswani2017attention}. It can be categorized into the soft attention. The non-local block, introduced in ~\cite{DBLP:journals/corr/abs-1711-07971}, is highly related to the self-attention module, but captures long-range dependencies in space-time dimension in images and videos. \cite{yue2018compact,DBLP:conf/cvpr/ZhengFZL19} further explore the non-local like ideas in fine-grained classification.

In contrast to these self-attention based methods,
we exploit the interactions between channels to discover
the  channel-wise complementary information rather than mining the closely related channels.
Moreover, we further propose a contrastive channel interaction module to model cross-sample channel interactions.

\textbf{Metric learning.} Deep metric learning aims to learn a feature embedding for better measuring the similarities between image pairs, i.e., the distance of positive pairs are encouraged  to be closer and the negative pairs are pushed away from each other. It has been widely used in various domains such as face verification~\cite{DBLP:conf/cvpr/HuLT14,DBLP:conf/cvpr/SchroffKP15}, image retrieval~\cite{DBLP:conf/cvpr/WangSLRWPCW14}, person re-id~\cite{DBLP:conf/cvpr/ChenCZH17,DBLP:conf/eccv/VariorHW16}, etc.
Compared with softmax loss used in conventional classification networks, metric learning can embed the samples into a low-dimensional space capturing high intra-class variance, which is more suitable for fine-grained image categorization~\cite{DBLP:conf/cvpr/CuiZLB16}.
Recent work of MAMC~\cite{DBLP:conf/eccv/SunYZD18} adopts metric learning to compute the rich correlations between object parts, which inspired the current work.
But our major differences lie in two aspects: 1) MAMC utilizes two attention branches to compute the features of two different part, while we model the interplay between different channels \emph{explicitly} to extract the discriminative features; 2) a novel contrastive channel interaction module is proposed in our networks to emphasize the differences between contrastive samples.
\section{Methodology}

In this section, we present our proposed channel interaction network (CIN) for fine-grained image categorization, as illustrated in  Figure \ref{approach:whole}.
Given an image pair, the two images are first processed by a shared backbone, e.g., ResNet-50~\cite{DBLP:conf/cvpr/HeZRS16}, generating a pair of convolutional feature maps.
To compute channel-wise complementary information for each channel on the feature maps, a self-channel interaction (SCI) module is designed to model the correlations between different channels. Then we aggregate the discriminate features from the original feature maps and the complementary information jointly.
%
Finally, a contrastive channel interaction (CCI) module is designed with a contrastive loss to model the channel-wise relationships between two images. Compared with existing methods, our proposed CIN can be trained end-to-end in one stage, and also is readily applicable to other convolution neural networks.

\subsection{Self-Channel Interaction}
\label{sec:sa}

Being aware of the rich knowledge encoded in the feature channels, as shown in Figure \ref{fig:introduction}, we would like to explore the interaction between various channels.
Recent work~\cite{DBLP:journals/corr/abs-1709-01507,DBLP:conf/eccv/SunYZD18} tends to highlight the most distinct feature channels.
However, only focusing on the most discriminate channels might not fully explore the rich information from all channels.
Indeed, most of the channels are complementary to each other. We attempt to compute the channel-wise relationships to extract such complementary clues, and then encode them into the original features for fine-grained classification.
Thus we propose a simple yet effective self-channel interaction (SCI) module to achieve such ability, as shown in Figure \ref{approach:whole}.

Given an image $I$, let $X' \in \mathbb{R}^{w \times h  \times c}$ denote the input feature maps processed by the backbone, where $w$, $h$ and $c$ indicate the height, width and the number of channels.
We first reshape the input feature maps $X'$ to $X \in \mathbb{R}^{c \times l}, l =w \times h$. Then the output of SCI is computed as:
\begin{equation}\label{eq:atten-output}
Y = WX \in \mathbb{R}^{c \times l},
\end{equation}
where $W \in  \mathbb{R}^{c \times c}$ denotes the SCI weight matrix, which can be computed as follows.
Firstly, we perform a bilinear operation between $X$ and $X^\top$, obtaining a bilinear matrix, $XX^\top$. Then we add a minus sign to it and exploit a softmax function to get the weight matrix:
\begin{equation}\label{eq:atten-input}
W_{ij} = \frac{exp(-{XX^\top}_{ij})}{\sum_{k=1}^{c}exp(-{XX^\top}_{ik})},
\end{equation}
where $\sum_{k=1}^cW_{ik} = 1$.
It is worth noting that $Y_i$ (the $i^{th}$ channel of the resulting features $Y$) is the computed interaction between $X_i$ and all the channels of $X$, i.e., $Y_i = W_{i1}X_1 + \dots + W_{ic}X_c$.

\begin{figure}[t]
\begin{center}
\includegraphics[width=0.48\textwidth]{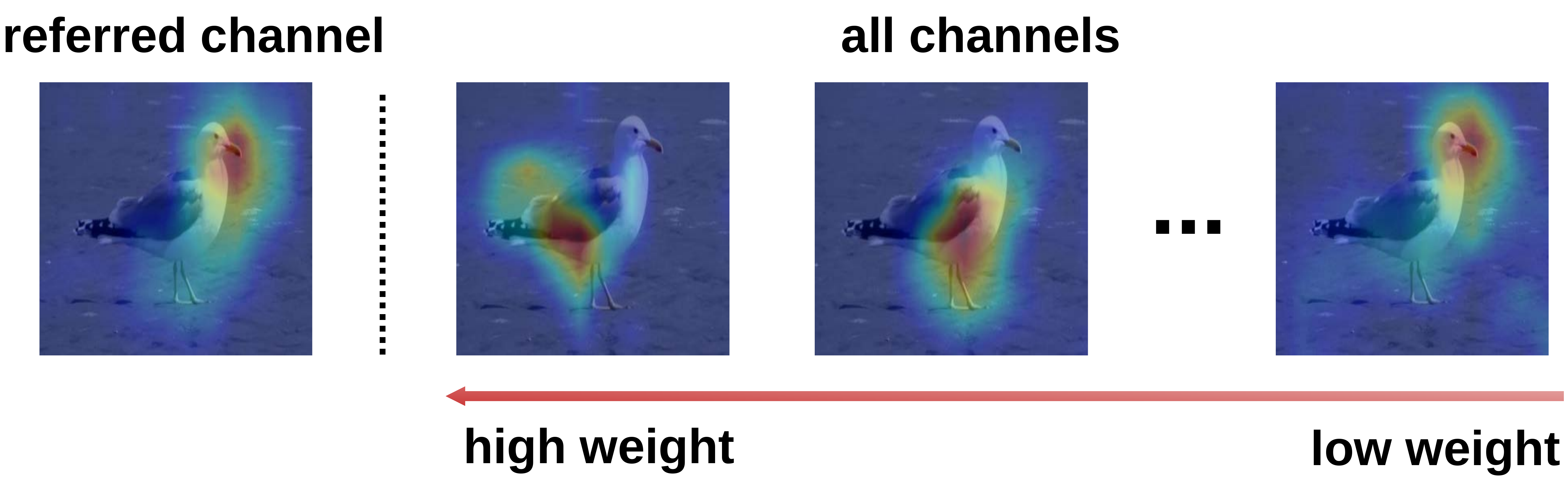}
\end{center}
\caption{An example of the relationship between the referred channel with all the channels in SCI module.}
\label{approach:sci}
\end{figure}

According to the definition of $W$, the channels with larger weights tend to be semantically complementary with $X_i$, as illustrated in Figure \ref{approach:sci}. The referred channel $X_i$ focuses on the head part, thus the channels highlighting the complementary parts, like wings and feet, have larger weights, while the channel with head part emphasized has a smaller weight. As the resulting features $Y$ may discard some information from the original features, we aggregate the  discriminate features ($Z$) from both the generated features and the original ones:
\begin{equation}\label{eq:sci-res}
Z = \phi(Y)+X,
\end{equation}
where $\phi$ denotes a $3 \times 3$ convolutional layer.

\textbf{Discussions.} It is worth noting that our SCI module can be formalized as the non-local like operation described in~\cite{DBLP:journals/corr/abs-1711-07971}:
\begin{equation}\label{eq:atten-no-local}
Y = f(X,X)g(X),
\end{equation}
where $f(X,X) = softmax(-XX^\top) \in \mathbb{R}^{c \times c}$, and $g(X) = X \in \mathbb{R}^{c \times l}$. Unlike the original non-local block considering the interactions in spatial dimension, our module focuses on channel dimension.
More importantly, the non-local operation tends to exploit the positive correlations between spatial positions, while our SCI module focuses on the negative correlations, which enables our model to discover the semantically complementary channel information.
The non-local operation is similar to the SE-Inception module~\cite{DBLP:journals/corr/abs-1709-01507}. They highlight the discriminative features but do not make full use of the complementary clues, which are better explored by our SCI module to enhance the channel-wise features, as shown in Figure \ref{exp:contrastive}.

Our method is related to CGNL~\cite{yue2018compact} and TSAN~\cite{DBLP:conf/cvpr/ZhengFZL19} which also compute the channel correlations, but has clear
distinctions on measuring the correlations:
1) CGNL and TSAN explore positive channel interaction while our CIN focuses on negative channel interaction;
2) CGNL takes spatial correlation into account, and further posits a low-rank Hadamard product;
3) In TSAN, the authors further proposed an adaptive image sampling mechanism to enhance the detailed information
and applied knowledge distilling to extract the learned details.
4) Beside computing the channel-wise relationship within an image, a key distinction of our method is to  further apply metric learning to model the channel interplay between samples.

\subsection{Contrastive Channel Interaction}
\label{sec:cca}
SCI module is able to compute meaningful discriminate features. A straightforward approach is to directly feed the features for classification, e.g., by using a softmax classifier as the most popular choice.
However, a vanilla classifier usually fails to capture the subtle differences present for fine-grained classification~\cite{DBLP:conf/cvpr/CuiZLB16}. To mitigate this problem, MAMC~\cite{DBLP:conf/eccv/SunYZD18} was recently proposed to enforce the correlations between different object parts. It introduces multi-attention multi-class constraints by using a metric learning technology, which inspired the current work.
We employ deep metric learning to compute rich cross-sample channel-wise correlations by introducing contrastive constraints to the features enhanced by SCI.

To model this interaction between two images $I_A$ and $I_B$, a natural idea is to impose the contrastive constraints on the features  $Z_A$ and $Z_B$ enhanced by SCI, and then measure their similarity. However, traditional deep metric learning approaches project an image into a fixed point in the learned embedding space. As a result, such a general representation often fails to capture the subtle differences between two images. In contrast, we attempt to learn the interactions between two images in a dynamic manner where the channels are emphasized by comparing to the feature channels computed from the contrastive image.

Consequently, we propose a contrastive channel interaction (CCI) module to compute such relationships between two images. As illustrated in Figure \ref{approach:whole}, we argue that a simple subtraction operation between the SCI weight matrices of image $I_A$ and $I_B$ might extract such mutual information, and generate CCI weight matrices $W_{AB}$ and $W_{BA}$:
\begin{equation}\label{eq:cci-matrix}
W_{AB} = |W_A - \eta W_B|,\\
W_{BA} = |W_B - \gamma W_A| ,
\end{equation}
where $\eta$ and $\gamma$ are the weights learned by $[Y_A,Y_B]$ and $[Y_B,Y_A]$ through a FC layer $\psi$, i.e., $\eta = \psi([Y_A,Y_B]), \gamma = \psi([Y_B,Y_A])$, and $||$ denotes the absolute value.
The two weights indicate the amount of correlated information considered dynamically by the image to better distinguish itself from the compared one.
We use a subtraction operation to compute the interaction. We also tried other operations like addition, multiplication, or concatenation, with slightly lower performance obtained. By subtraction,
the CCI weight matrices suppress the commonality and highlight the distinct channel relationships between the two images.

Then similar to SCI module, the CCI weight matrices $W_{AB}$ and $W_{BA}$ are applied to the features $X_A$ and $X_B$ as:
\begin{equation}\label{eq:atten-input}
Z_{A}' = \phi(Y_{A}')+X_A,  \\
Z_{B}' = \phi(Y_{B}')+X_A,
\end{equation}
where $Y_{A}' = W_{AB}X_A$ and $Y_{B}' = W_{BA}X_B$.

Finally, a contrastive loss~\cite{hadsell2006dimensionality} is applied to the features computed by the CCI module which aims to push the samples of different classes away while pulling the positive image pairs close.
Suppose each batch contains N image pairs, i.e., 2N images.
The contrastive loss is defined as follows:
\begin{equation}\label{eq:atten-input}
L_{cont} = \frac{1}{N}\sum_{A,B}\ell(Z_{A}', Z_{B}').
\end{equation}
Beyond the contrastive loss, a triplet loss~\cite{DBLP:conf/cvpr/SchroffKP15} and other losses of metric learning can be used in our framework as well. The reason we choose the contrastive loss is that it is simple, and perform well in metric learning and face verification~\cite{DBLP:conf/cvpr/TaigmanYRW14,hadsell2006dimensionality}. We also tried to use a triplet loss in CCI, but did not improve the performance. Specifically, $\ell$ is defined as follows:
\begin{equation}\label{eq:contras_loss}
\ell =
\begin{cases}
||h(Z_{A}')-h(Z_{B}')||^2,  & \mbox{if } y_{AB} = 1 \\
max(0, \beta - ||h(Z_{A}')-h(Z_{B}')||)^2, & \mbox{if } y_{AB} = 0
\end{cases}
\end{equation}
where
$\beta$ is a predefined margin and $||\cdot||$ denotes the Euclidean distance, $h$ is a fully-connected layer projecting features into an $r$-dimension space, i.e. $H(Z) \in \mathbb{R}^{r}$.
$r$ is set to 512 in our experiments.
Here, $y_{AB}$ indicates whether the label of an image pair is the same or not, i.e., $y_{AB}=1$ denotes image $I_A$ and image $I_B$ come from the same class, while $y_{AB}=0$ means a negative pair.

Moreover, we use a softmax loss for classification based on the predictions that are generated by the features $Z$ using SCI.
We denote the softmax loss as $L_{soft}$.
The total loss $L_{total}$ of our framework is defined as follows:
\begin{equation}\label{eq:loss}
L_{total} = L_{soft} + \alpha \cdot L_{cont},
\end{equation}
where $\alpha$ is a hyper-parameter.
We use the stochastic gradient method to optimize $L_{total}$.
Note that only SCI module is used in inference, with only a single image required.

\section{Experiments}
We report the experimental results, and compare our method with the state-of-the-art approaches.

\subsection{Datasets and Baselines}
\label{sec:data-basline}

\noindent\textbf{Datasets.} We employ three publicly available datasets in our experiments:
(1) CUB-200-2011~\cite{wah2011caltech} with 11,788 images from 200 wild bird species,
(2) Stanford Cars~\cite{krause20133d} including 16,185 images over 196 classes,
and
(3) FGVC Aircraft~\cite{DBLP:journals/corr/MajiRKBV13} containing 196 classes about 10,000 images.

\noindent\textbf{Baselines.}
In the experiments, we compare our CIN with 10 methods described as follows.
The first four methods can be trained in one stage.
(1) MAMC~\cite{DBLP:conf/eccv/SunYZD18}: applying multi-attention multi-class constraints to enforce the correlations among different parts of objects.
(2) CGNL~\cite{yue2018compact}: capturing the dependencies between positions across channels by non-local operation to classify.
(3) HBP~\cite{yu2018hierarchical}: hierarchical bilinear pooling framework integrating multiple cross-layer bilinear features.
(4) iSQRT-COV~\cite{yu2018hierarchical}: using an iterative matrix square root normalization to do covariance pooling.
(5) RA-CNN~\cite{fu2017look}: recursively learning discriminative region attention and region-based feature representation at multiple scales.
(6) Boost-CNN~\cite{DBLP:conf/bmvc/MoghimiBSYVL16}: a new boosting strategy to assemble weak classifiers for better performance.
(7) DT-RAM~\cite{li2017dynamic}: a dynamic computational time model with reinforcement learning for recurrent visual attention.
(8) MA-CNN~\cite{zheng2017learning}: multi-attention convolutional network including convolution, channel grouping and part classification sub-networks.
(9) DFL-CNN~\cite{wang2018learning}: capturing class-specific discriminative patches by learning a bank of convolutional filters. The performance might be unstable due to the complex layer initialization using k-means.
(10) NTS~\cite{yang2018learning}: effectively localizing informative regions with self-supervision mechanism.

Notice that we do not compare our method  with the approaches which require additional information, such as SJS \cite{ge2017borrowing}, HS-Net~\cite{lam2017fine}, and  HSE \cite{chen2018fine}.

\subsection{Implementation Details}
In all our experiments, we use ResNet-50 and ResNet-101 as our base networks. We remove the last pooling layer and fully-connected layer, and then fine-tune the networks pretrained on ImageNet~\cite{DBLP:journals/ijcv/RussakovskyDSKS15}. The input image size is 448 $\times$ 448 as most state-of-the-art fine-grained categorization approaches. By following that of NTS~\cite{yang2018learning}, we implement data augmentation including random cropping
and horizontal flipping during training. Only center cropping is involved in inference.

The model is trained for 100 epochs with SGD for all datasets, and the base learning rate is set to 0.001, which annealed by 0.5 every 20 epochs.
we use a batch size of 20 and ensure that each batch contains 4 categories with 5 images in each category. And then, we randomly split
these 20 images into 10 image pairs. We have tried to use
all the $O(n^2)$ pairs or apply hard negative mining, which
hurt the performance and consume more memory.
The weight decay is set to $2 \times 10^{-4}$. $\beta$ in Equation \ref{eq:contras_loss} is set to 0.5 empirically. and $\alpha$ in Equation \ref{eq:loss} is set to 2.0. Top-1 accuracy is used as the evaluation metric. We use PyTorch to implement our method.

\subsection{Ablation Analysis}
\label{sec:ablation}

We conduct ablation studies in order to better understand the impact of each component to our approach.
The performance and efficiency are compared in in Table \ref{tab:cca-ab}.
We use ResNet-50 and ResNet-101~\cite{DBLP:conf/cvpr/HeZRS16} as our backbone.

\begin{table}
\small
\begin{center}
\scalebox{0.93}{
\begin{tabular}{|l|c|c|c|}
\hline
Method & 1-Stage & ACC & Time\\
\hline\hline
VGG-19 & $\surd$ & 80.2\% & 22.1\\
ResNet-50 & $\surd$ & 84.9\% & 12.5\\
ResNet-101 & $\surd$ & 85.4\% & 22.4\\
ResNet-50 + SE & $\surd$ & 85.7\% & 14.0\\
ResNet-50 + Pos-SCI & $\surd$ & 86.1\% & 17.2\\
ResNet-50 + Non-local & $\surd$ & 86.6\% & 14.2\\
ResNet-50 + MAMC~\cite{DBLP:conf/eccv/SunYZD18} & $\surd$ & 86.3\% & 14.8\\
ResNet-50 + CGNL~\cite{yue2018compact} & $\surd$ & 87.0\% & 15.0\\
ResNet-50 + SCI & $\surd$ & 87.1\% & 17.2\\
ResNet-50 + SCI + Cont & $\surd$ & 87.2\% & 17.2\\
ResNet-50 + NTS~\cite{yang2018learning} & $\times$ & 87.5\% & 23.6\\
ResNet-50 + CIN & $\surd$ & 87.5\% & 17.2\\
ResNet-101 + CIN & $\surd$ & \textbf{88.1\%} &27.2\\
\hline
\end{tabular}}
\end{center}
\caption{Ablation studies of our network on CUB-200-2011. CIN consists of SCI and CCI. Time unit is ms.}
\label{tab:cca-ab}
\end{table}

\begin{table*}[t]
\begin{center}
\begin{tabular}{|l|c|c|c|c|}
\hline
Method  & 1-Stage & Acc(CUB)  & Acc(FGVC) & Acc(Stanford Cars)\\
\hline\hline
MAMC~\cite{DBLP:conf/eccv/SunYZD18} & $\surd$ & 86.5\% & - & 93.0\%\\
CGNL~\cite{yu2018hierarchical}  & $\surd$ & 87.0\%& - & - \\
HBP~\cite{yu2018hierarchical}  & $\surd$ & 87.1\%& 90.3\% & 93.7\%\\
iSQRT-COV(8k)~\cite{DBLP:conf/cvpr/LiXWG18} & $\surd$ & 87.3\% & 89.5\% & 91.7\%\\
RA-CNN~\cite{fu2017look}  & $\times$ & 85.3\% & - & 92.5\%\\
Boost-CNN~\cite{DBLP:conf/bmvc/MoghimiBSYVL16}  & $\times$ & 85.6\% & 88.5\% & 92.6\%\\
DT-RAM~\cite{li2017dynamic} & $\times$ & 86.0\% & - & 93.1\%\\
MA-CNN~\cite{zheng2017learning}& $\times$ & 86.5\% & 89.9\% & 92.8\%\\
DFL-CNN~\cite{wang2018learning} & $\times$ & 87.4\% & 92.0\% & 93.8\%\\
NTS~\cite{yang2018learning} & $\times$ & 87.5\% & 91.4\% & 93.9\%\\
\hline
CIN (ResNet-50) & $\surd$ & 87.5\% & 92.6\% & 94.1\%\\
CIN (ResNet-101) & $\surd$ & \textbf{88.1\%} & \textbf{92.8\%}& \textbf{94.5\%}\\
\hline
\end{tabular}
\end{center}
\caption{Comparison results on CUB-200-2011, FGVC Aircraft and Stanford Cars.}
\label{tab:cub}
\end{table*}

\textbf{SCI Module.}
SCI mines complementary channels through exploring channel interactions, contributing to learning more discriminative features.
As illustrated in Table \ref{tab:cca-ab}, compared with ResNet-50 alone (84.9\%), by merely adding the SCI module, ResNet-50 + SCI obtains a performance improvement of 2.2\%. Moreover, switching the interaction module from SCI to SE module leads to a significant performance drop (87.1\% vs. 85.7\%).
SE module only focuses on the most discriminative features and ignores others, while our SCI module utilizes the complementary channel knowledge to enhance all the features.
Compared to the Non-local block and ResNet-50+Pos-SCI (SCI weight matrix $W$ without the negative sign) which model the positive space and channel-wise information respectively, our SCI module obtains better performance.
 Notice that our SCI also outperforms
CGNL~\cite{yue2018compact} (87.0\%) which models the correlations between the positions of all channels.
Indeed, the channel information explored in our SCI module is involved in CGNL as well.
The major difference about the channel information lies in that our SCI exploits the negative interplay to find the channel-wise complementary  information, while CGNL does not fully explore such information and computes the positive interaction to capture the closely related clues.
These results demonstrate that: 1) for fine-grained image classification, the information contained in the channel dimension is as powerful as complicated modeling across all dimensions; 2) finding the complementary channel clues can take full advantage of the channel interaction comparing with discovering the closely related channel information.

\textbf{CCI Module.} We further investigate the effectiveness of the proposed CCI module.
Table \ref{tab:cca-ab} shows that the CCI module (ResNet-50+SCI+CCI) provides 0.4\% performance improvement compared to the method without a contrastive loss (ResNet-50+SCI).
To further demonstrate the characteristics of the contrastive channel attention module, we consider the approach (ResNet-50+SCI+Cont) which explicitly applies a contrastive loss to the features computed by SCI module, i.e., $\eta=0$ and $\gamma=0$ in Equation \ref{eq:cci-matrix}.
As presented in Table \ref{tab:cca-ab}, ResNet-50+SCI+Cont obtains a limited improvement with ResNet-50+SCI (87.2\% vs. 87.1\%).
The reason might be that the common contrastive loss uses the same features of an image compared to any other image, which might reduce it is ability to focus on the distinct differences between two images, while our CCI module is capable of highlighting of the different regions.
The results confirm that our CCI module has strong capability for modeling the relationship between two images.

\textbf{Time cost.} We report our inference time on a Nvidia TI-
TAN XP GPU with PyTorch implementation. As shown in Table \ref{tab:cca-ab}, CIN introduces an overhead that is much smaller than that of two-stage methods (ResNet-50+NTS), and is comparable to the other one-stage approaches.

\subsection{Comparison with State-of-the-art}
In this section, we compare our proposed network (CIN) with the state-of-the-art methods on the three publicly available datasets.

\textbf{CUB-200-2011.} Table \ref{tab:cub} presents the classification results of CIN and the state-of-the-art methods.
First, the accuracy of our proposed CIN is higher than all existing methods. Even with ResNet-50, our method achieves comparable result with NTS~\cite{yang2018learning}. However, NTS requires multiple stages for learning discriminative regions, resulting in more expensive cost on both time and space.
Compared with the best one-stage method iSQRT-COV (8k)~\cite{DBLP:conf/cvpr/LiXWG18}, our method outperforms it by 0.2\%. Note that the feature dimension of our method (2k) is significantly lower than iSQRT-COV (8k).
Moreover, our method improves HBP~\cite{yu2018hierarchical} by 1.0\%. The reason might be that HBP ignores the interaction between samples. It is notable that the backbone of HBP is VGG~\cite{simonyan2014very}, while the accuracy of CIN with the same backbone is 85.6\%. We have tried to implement HBP and found it does not work well with ResNet. DFL-
CNN~\cite{wang2018learning} achieves the best results on CUB (87.4\%) with ResNet-50 backbone while ResNet-50+CIN achieves a higher
accuracy with only one stage. As shown in ~\cite{wang2018learning}, ResNet
does not always outperform VGG.

\begin{table}
\begin{center}
\begin{tabular}{|l|c|c|c|}
\hline
Dataset & CIN & NTS & NTS+CIN \\
\hline\hline
CUB-200-2011 & 87.5\% & 87.5\% & \textbf{88.3\%}\\
FGVC Aircraft & 92.6\% & 91.4\% & \textbf{93.3\%}\\
Stanford Cars & 94.1\% & 93.9\% & \textbf{94.4\%}\\
\hline
\end{tabular}
\end{center}
\caption{Combined our network with NTS.}
\label{tab:nts-ab}
\end{table}

\begin{figure}[t]
\begin{center}
\includegraphics[width=0.48\textwidth]{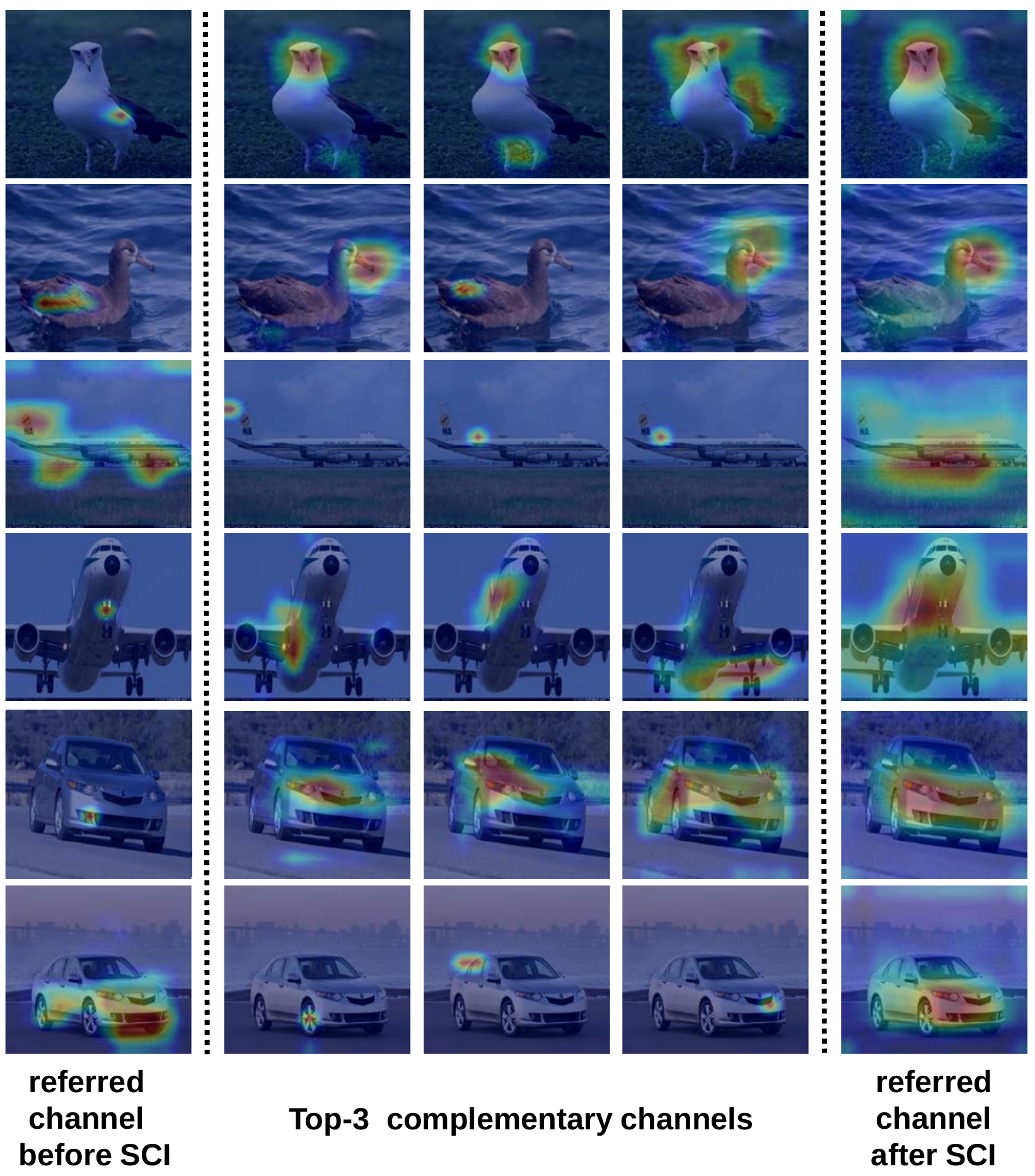}
\end{center}
\caption{Visualization of channel activations before and after SCI moudle on CUB, Cars and Aircraft.}
\label{exp:attention}
\end{figure}

\begin{figure}[t]
\begin{center}
\includegraphics[width=0.48\textwidth]{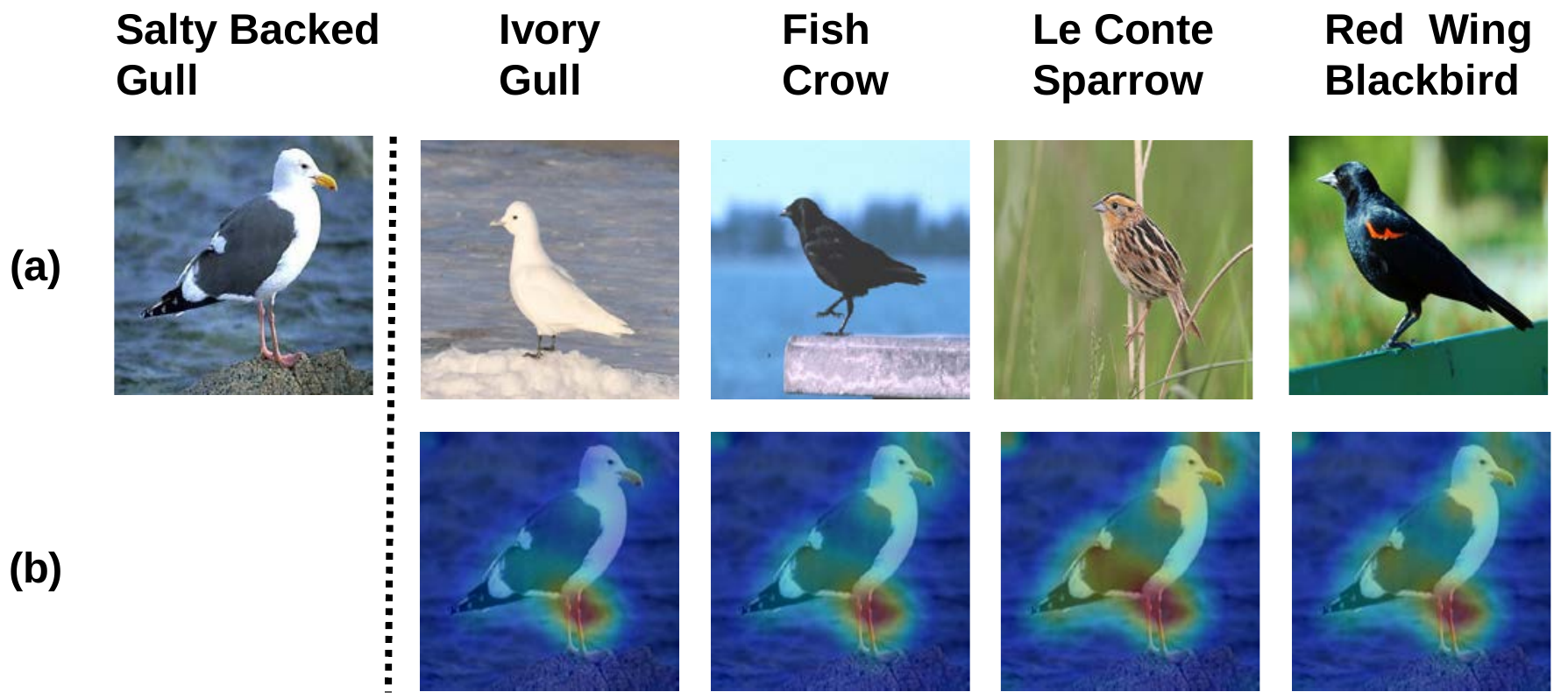}
\end{center}
\caption{Visualization on the results of CCI module on CUB. (a) the original images; (b) the feature maps by CCI. It can be seen that different regions are highlighted conditioned on different image pairs.}
\label{exp:contrastive}
\end{figure}

\textbf{FGVC Aircraft.} Table \ref{tab:cub} reports the performance on FGVC Aircraft dataset.
DFL-CNN achieves the highest accuracy of 92.0\%, outperforming  NTS with 91.4\%.
Our method has a clearly higher accuracy than existing methods even with the backbone ResNet-50. The excellent results further confirm the superiority of our method. It is worth noting that the accuracy on FGVC Aircraft is generally higher than that of CUB-200-2011, because images of CUB-200-2011 contain much more label noises (4.4\% as reported in ~\cite{van2015building}) and class-irrelevant background, while images in FGVC Aircraft have relatively clean background, and airplanes often occupy a large portion of the image.

\textbf{Stanford Cars.} To verify the generalization ability of the proposed method.
 We further evaluate it on another real-world dataset, the Stanford Cars. Table \ref{tab:cub} presents the performance of our method with the state-of-arts. Generally, the results are consistent with those of the previous two datasets. Again, the proposed CIN can achieve the highest accuracy compared with the state-of-arts.

\textbf{Combined with NTS.} Furthermore, our module is general and flexible, and it can be readily integrated into other framework to improve the performance. In this experiment, we combine our module with the latest state-of-the-art method NTS~\cite{yang2018learning}, which is a two-stage framework by leveraging a region proposal networks to localize discriminative parts with weakly-supervised learning. We integrate the SCI module at the end of the feature extractor networks. As NTS will discover multiple regions out of sequence, thus we only apply the CCI to the whole feature stream. Table \ref{tab:nts-ab} shows the performance of our method combined with NTS (NTS+CIN). As can be found, NTS+CIN achieves \emph{consistent} performance improvements on all the three publicly available datasets compared with either NTS or CIN alone. The results further demonstrate the strong capability of our module. We expect that our CIN network can improve the performance on various computer vision tasks when simply plugged into existing framework.

\subsection{Qualitative Visualization}
To better understand the intra- and inter-image channel interactions modeled by CIN, we visualize the channel correlations and neural activations in our SCI and CCI module. Figure \ref{exp:attention} shows the visualization of SCI module for images from three different datasets. Column 1 presents the activations of a randomly selected channel (assuming it is the $i^{th}$ channel) before SCI. Column 2 to Column 4 are the three most complementary channels to it.
In other words, these three channels correspond to the ones that have the largest values in the $i^{th}$ row of SCI matrix $W$ defined in Equation \ref{eq:atten-input}. The last column represents $Y_i$, which is the $i^{th}$ channel after SCI. We find that, for a referred channel, the top-3 complementary channels tend to capture different semantic.
For instance, in the first example of Figure \ref{exp:attention}, the referred channel has a strong activation around wings, and its complementary channels focus more on head and tail regions. As a result, the attention feature channels are enhanced by this complementary information and activates also on other discriminative parts.
Note that after our SCI module, the activations span most of the object parts, which indicates that SCI effectively models the interactions among different channels, and combine their complementary but discriminative parts to produce more informative features.

Figure \ref{exp:contrastive} visualizes the results of our CCI module on CUB-200-2011 dataset. Line 2 shows the contrastive attention activations by averaging all feature maps after CCI across channels. ``Salty Black Gull" and ``Ivory Cull" have similar heads, and their features after CCI have weaker responses to the head. While comparing with ``Fish Crow", the activations near the head becomes stronger. For the other two bird species, their appearance differences are huge and the CCI module
provides strong responses to the whole body part.
This result suggests that our proposal CCI module can focus on the key distinctions by modeling the interactions of channels between image pairs.

\section{Conclusion}
We have presented a new channel interaction network (CIN) for fine-grained image categorization. Our network first learns complementary channel information by a self-channel interaction (SCI) module taking the relationships between channels into account. It encourages to pull positive pairs closer while pushing negative pairs away via a contrastive channel interaction (CCI) module, which exploits channel correlations between samples. The proposed network can be trained end-to-end in one stage requiring no bounding box/part annotations. Extensive experiments demonstrate that CIN can achieve superior performance compared to the state-of-the-art approaches.

\begin{quote}
\begin{small}
\bibliographystyle{aaai}\bibliography{egbib}
\end{small}
\end{quote}

\end{document}